\documentclass[letterpaper]{article}
\usepackage[preprint,ccby]{aaai2027}
\usepackage[hyphens]{url}
\usepackage{graphicx}
\usepackage{natbib}
\usepackage{caption}
\usepackage{booktabs}
\usepackage{amsmath}
\title{Signal or Spurious Cue? A Randomized Audit of Survey-Country Metadata in LLM Social Inference}
\author{Yifan Lyu, Xinran Li, Jiaqi Qiao, and Xiujuan Xu}
\affiliations{Dalian University of Technology, China}

\begin{document}

\maketitle

\begin{abstract}
Survey-country metadata can improve an LLM's forecast of an individual response when informative, yet the same cue may redirect the forecast when assigned at random. A within-record audit tests whether disclosing a random label's uniform, record-independent origin reduces its country-directed uptake, and whether verified survey country lowers held-out Brier loss. Independent population anchors and recorded human answers measure direction and consequence across five fixed API models, six countries, and seven development-selected targets. In the primary post-review 72-record panel, opaque and disclosed-random labels each produced country-direction shifts of 0.214. Paired attenuation was 0.0003 (95\% CI $[-0.0157, 0.0166]$). Verified country reduced Brier loss by 0.040 (95\% CI $[0.024, 0.056]$), while random-label regret included zero. A non-overlapping mixed-coverage consistency panel retained positive disclosed-random movement and verified utility, while attenuation remained uncertain. On the selected targets, verified metadata was useful in both panels, but disclosure did not reliably attenuate random-label uptake. PROV-FORECAST contains 14,400 paired item-level probability distributions from the corrected panel.
\end{abstract}

\section{Introduction}

\begin{figure}[t]
\centering
\includegraphics{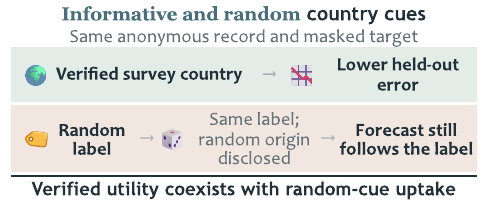}
\caption{Directional movement alone does not establish predictive harm.}
\label{fig:intro-tension}
\end{figure}

Group metadata can improve individual forecasts when it carries population information, but its wording can also redirect a forecast. Distinguishing information from cue-driven movement matters when LLMs infer unseen responses from sparse individual evidence. Figure~\ref{fig:intro-tension} separates direction from predictive consequence.

Survey evaluations compare model outputs with group opinion distributions \citep{santurkar2023opinions,durmus2024global}. Persona prompts test how group descriptions redirect responses \citep{mukherjee2024cultural,lutz2025prompt,col2026geographic}, while individual-response methods forecast held-out answers from observed stances \citep{malone2025beliefs,anonymous2026individual}.

These methods establish aggregate gaps, prompt sensitivity, or predictive value.

They do not identify whether the same group label has a different effect when its provenance changes, although source validity determines whether the field should inform a forecast \citep{locksley1980stereotypes,kunda1996impressions}.

The identification challenge has two parts. A provenance contrast must hold the displayed country and individual evidence fixed, and any induced movement must be separated from predictive benefit or harm. The experiment therefore pairs an opaque random country with the same label disclosed as uniformly assigned and record-independent, then compares a verified survey-country policy with no country field. Independent population distributions measure whether forecasts move in the displayed country's direction. Held-out Brier loss measures whether the changed forecasts better predict the recorded answer.

\textbf{RQ1} asks whether disclosure reduces country-directed uptake of the random label. \textbf{RQ2} asks whether verified survey country lowers held-out loss. The evaluation spans five models, six countries, and seven development-selected targets. It combines a 504-record panel with target-specific valid coverage and a non-overlapping 72-record panel that applies the corrected protocol to every target. The design contributes a same-label provenance contrast, an external direction measure without an LLM judge, and a consequence measure separating cue uptake from predictive utility or harm. It further contributes \emph{PROV-FORECAST}, a provenance-structured corpus of 14,400 item-level probability distributions from the corrected panel, aligned across five API models and four paired conditions.

The submitted archive preserves paired distributions and verifies its released vectors, while held-out Brier recomputation requires lawful source-data access. RQ2 evaluates the complete verified-metadata policy because field presence, identity, and wording change together.

\section{Related Work}

\paragraph{Survey opinion and response prediction.}
OpinionQA and GlobalOpinionQA expose representation gaps by comparing model answers with survey distributions \citep{santurkar2023opinions,durmus2024global}. Individual-response work predicts held-out answers from prior stances and tests whether demographic fields add information \citep{malone2025beliefs,anonymous2026individual}. Aggregate agreement asks whose opinions a model resembles. Individual forecasting asks whether an unseen response can be predicted. Randomising the metadata channel separates the displayed-country effect from the record's survey country.

\paragraph{Persona and cultural conditioning.}
Persona selection, value reasoning, and sociodemographic prompting can steer models towards specified populations \citep{long2025aligning,tehenan2025mpta,lutz2025prompt,tan2026persona}. Cultural and geographic prompts can affect tasks with little evident cue relevance \citep{mukherjee2024cultural,col2026geographic}, while cultural and personal alignment can conflict \citep{borah2026norms,wedgwood2026personalization}. These studies treat a persona as a desired perspective or observed attribute. The present comparison varies the provenance of one randomly assigned label and adds a verified-country positive control.

Bias benchmarks and marked-persona tests show that group cues can alter model judgements and expose stereotyped associations \citep{parrish2022bbq,cheng2023marked}. The present outcome is narrower: it separates country-directed movement from held-out predictive loss rather than treating any group-conditioned difference as bias.

\paragraph{Source discernment and measurement validity.}
Information-discernment research tests whether models use a claim according to source and truth status \citep{ashkinaze2026discernment}. Survey-based LLM evaluation faces wording, translation, and construct-validity risks \citep{libovicky2026credibility}. Cross-national comparisons also require caution about measurement equivalence \citep{davidov2014equivalence}. The present setting attaches provenance to group metadata in a social inference task, evaluates label-induced movement against survey-derived directions, and scores prediction against human answers. The resulting evidence concerns this metadata protocol rather than a general account of source reasoning or a model's internal cultural beliefs.

Randomised metadata contrasts resemble counterfactual audits, but counterfactual fairness requires a causal fairness criterion not identified here \citep{kusner2017counterfactual}. Elicited probabilities can also be evaluated for calibration \citep{jiang2021calibration,tian2023calibration}. This study uses them only as forecasts scored by proper loss.

\section{Experimental Design}

The experiment follows four stages. Reference records define country-level response directions, development records select targets, evaluation records provide anonymous observed and held-out answers under balanced metadata assignments, and five API models produce paired forecasts that are scored locally. Table~\ref{tab:study-inventory} summarises the resulting design.

\subsection{Data and Splits}

The Joint EVS/WVS 2017--2022 Dataset v5.0 contains 156,658 harmonised respondent records and 231 joint variables \citep{evswvs2024joint}. Restricting the source to China 2018, France 2018, Great Britain 2018, Italy 2018, Jordan 2018, and the United States 2017 yields a 12,770-record study pool. These waves were selected for common-item coverage and adequate sample size before endpoint outcomes were analysed. Survey country denotes the country in which the survey data were collected. It is neither citizenship nor a claim about cultural identity.

The 12,770 records were assigned by a stable hash to reference, development, pilot, and evaluation splits containing 6,412, 3,782, 1,286, and 1,290 records. Reference records estimate weighted country distributions. Development records determine eligible evidence and target items without access to evaluation outcomes. Pilot records support output and protocol diagnostics. Evaluation records supply the masked answers used for scoring. No record crosses splits.

\subsection{Task Construction}

Each model prompt contains ten observed answers from one anonymous record. The target is a set of answers masked from that same record. Harmonised English wording and response options are used. Checklist indicators are rendered as standalone binary questions with ``No'' and ``Yes'' options aligned to their recorded values. The model returns one probability for each displayed option. The recorded human choice remains a single option. It becomes a one-hot vector only for scoring, so probabilistic model forecasts and deterministic survey answers are compared by a proper scoring rule rather than treated as the same response format \citep{brier1950verification,gneiting2007proper}.

Target selection used development data. A regularised multinomial baseline first predicted each answer from observed record evidence. Adding survey country defined incremental country value. A high-country-value target required positive lower 95\% bounds for both evidence gain and incremental country gain, with incremental country gain at least 5\% of evidence-only Brier loss. Seven targets met these rules after questions that made a displayed country a direct referent were excluded. Evaluation answers were not used for selection. The seven scored questions and three unscored fillers formed two fixed five-question batches. Filler outputs were retained but excluded from the reported estimates.

The seven high-value targets cover three neighbour-group indicators, two child-quality indicators, petition signing, and lawful peaceful demonstration. The stratum was selected because survey country had out-of-sample predictive value in development data, not because any LLM had already shown a metadata effect. This distinction prevents the target rule from mechanically producing either country-directed movement or verified-country utility in the evaluation records.

For example, a request may show the recorded importance of religion and mask whether the same record excludes drug addicts as neighbours. The model returns an option-aligned vector such as $[0.70,0.30]$. Evidence, wording, batch, and option order remain fixed across metadata conditions.

\subsection{Evaluation Panel and Models}

The evaluation uses 576 distinct records across two non-overlapping panels (Table~\ref{tab:study-inventory}). The 504-record panel retains all 504 records for two wording-unaffected targets. For the five wording-affected targets, it replaces the invalid forecasts with corrected forecasts from a balanced 72-record subset of those same 504 records. Targets receive equal macro weight, so the two targets with wider record coverage do not dominate the estimates. This panel therefore contains 504 distinct records but has target-specific valid coverage.

The separate 72-record panel contains 12 different records per survey country, selected by a frozen hash after excluding all 504 records in the other panel. Every record-target forecast uses the corrected, hash-bound protocol for all seven targets. The correction was implemented after review and is neither a prospective preregistration nor an independent replication. The panel had no prospective power guarantee, and repeated targets were not treated as independent records. The two panels address the same estimands under different coverage structures. The 504-record panel supplies broader record coverage, while the separate 72-record panel supplies uniform corrected-protocol coverage across targets.

Five fixed API models were evaluated: DeepSeek-V4-Pro, Gemini-3.6-Flash, Mistral-Medium-3.5, GPT-5.4, and Qwen3.7-Max-2026-06-08. Prompts, target batches, option order, and metadata assignments were paired across models. The lowest-variance available settings were used, with explicit reasoning disabled where exposed. No LLM judge was used.

\begin{table}[t]
\centering
\small
\setlength{\tabcolsep}{3.2pt}
\begin{tabular}{@{}lp{1.78in}@{}}
\toprule
Component & Frozen scope \\
\midrule
Source/splits & 156,658 records, including 12,770 in six waves and four disjoint hash splits \\
Evaluation records & 576 distinct records in two non-overlapping panels \\
504-record panel & 2 targets $\times$ 504 and 5 targets $\times$ 72 corrected within-panel records \\
Separate 72-record panel & 7 targets $\times$ 72 under one corrected protocol \\
Prompt evidence & 10 observed answers per record \\
Task frame & 7 scored targets, 3 fillers, 2 batches \\
Metadata policies & 4 ($M_0,M_O,M_R,M_V$) \\
API models & 5, with identifiers in Table~\ref{tab:model-results} \\
Separate-panel calls & 2,880 valid of 2,880 planned \\
\bottomrule
\end{tabular}
\caption{Study inventory. The 72 corrected records used for five targets in the 504-record panel belong to that panel. They differ from the separate 72-record panel, which has no record overlap with the 504. One API call contains a five-target batch.}
\label{tab:study-inventory}
\end{table}

Responses were schema-validated JSON probability vectors. All 2,880 planned calls for the separate 72-record panel were valid, with no missing cells or local normalisation. Raw responses were retained unchanged.

\begin{figure*}[t]
\centering
\includegraphics{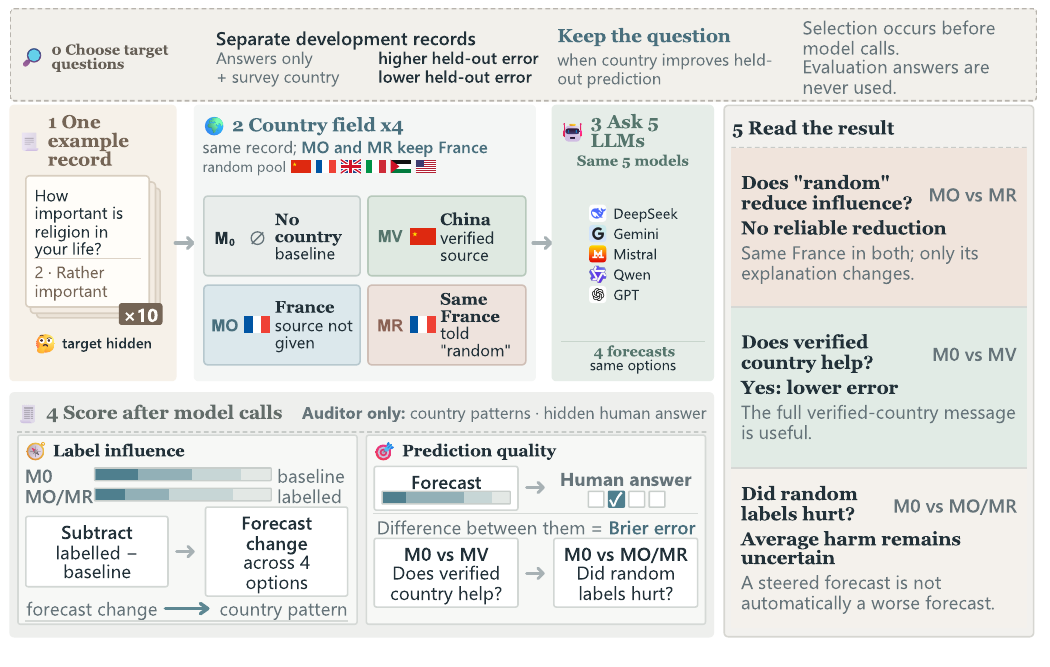}
\caption{Audit flow. The card and country fields are one schematic task, not a released record or output. Screening uses development-set prediction gain. Direction uses independent country anchors. Brier loss uses masked evaluation answers. Flags denote survey-country fields, not nationality or cultural identity.}
\label{fig:audit-design}
\end{figure*}

Figure~\ref{fig:audit-design} maps the model-visible experimental conditions to the analysis-only contrasts.

\subsection{Metadata Interventions}

For each record, $M_0$ contains no country field. $M_O$ displays a country label $T$ drawn uniformly from the six candidates and states that its selection probabilities and relationship to the source record are not provided. $M_R$ displays the same $T$ but states: ``Each candidate was selected with equal probability, independently of the source survey record.'' This paired contrast changes disclosed provenance while holding the country name fixed. $M_V$ displays the true survey country copied from the source record and identifies it as the country in which the data were collected.

Assignment was balanced within true survey country: each displayed country occurred twice among 12 records. One frozen assignment was shared across models, batches, and the $M_O/M_R$ pair, isolating the tested provenance sentence under the fixed protocol.

The random label matches source country for one sixth of records by chance. It is uninformative before its realised value is observed, not deliberately false.

$M_0$ to $M_V$ changes the complete verified-metadata policy, not truth alone. Models never see identifiers, anchors, weights, or target answers.

\section{Evaluation Measures and Inference}

\subsection{Country-Directed Movement}

Direction scoring asks whether a metadata-induced forecast change resembles the country contrast observed in independent human records. For target $j$ and survey country $c$, the reference split defines the weighted response distribution $H_{jc}$ and the six-country mean $\bar H_j$. Their difference
\begin{equation}
D_{jc}=H_{jc}-\bar H_j
\end{equation}
is an external direction, not a definition of culture. Let $p^{X}_{mij}$ be model $m$'s probability vector for record $i$, target $j$, and condition $X$, and let $\mathcal I_c$ contain evaluation records from survey country $c$. Define
\begin{align}
\Delta p^{X}_{mij} &= p^{X}_{mij}-p^{0}_{mij},\\
n^{X}_{mjc} &= \sum_{i\in\mathcal I_c} w_i
\langle\Delta p^{X}_{mij},D_{jT_i}\rangle,\\
q_{jc} &= \sum_{i\in\mathcal I_c} w_i
\lVert D_{jT_i}\rVert_2^2,\\
b^{X}_{mjc} &= n^{X}_{mjc}/q_{jc},\\
\beta_X &= \operatorname{MacroAvg}_{m,j,c} b^{X}_{mjc},
\quad X\in\{O,R\}.
\end{align}
Positive $\beta_X$ means that the forecast moved along the displayed country's reference contrast. At the cell level, a coefficient of one is a projection equal to one full reference-country contrast. The reported macro-average is on that projection scale. Thus $\beta=0.214$ is neither an accuracy nor a claim that 21.4\% of records changed. Direction does not by itself imply greater accuracy, harm, stereotyping, or a corresponding internal belief. The disclosure attenuation estimand is $A=\beta_O-\beta_R$.

\subsection{Held-Out Consequence}

Consequence scoring asks whether the changed forecast better predicts the masked human option. For one-hot answer $y$, multiclass Brier loss is $L(p,y)=\sum_k(p_k-y_k)^2$. Lower loss is better. Let $\mathcal M[\cdot]$ denote the same macro-average over model, item, and survey country. Verified-country utility and random-label regret are
\begin{align}
U_V &= \mathcal M[L(p^0,y)-L(p^V,y)],\\
R_X &= \mathcal M[L(p^X,y)-L(p^0,y)],\quad X\in\{O,R\}.
\end{align}
For a one-hot outcome, this multiclass Brier loss ranges from zero to two. Positive $U_V$ denotes an absolute loss reduction from the verified-country policy; positive $R_X$ denotes harm from a random label.

\begin{table}[t]
\centering
\small
\setlength{\tabcolsep}{1.2pt}
\begin{tabular}{@{}lcc@{}}
\toprule
Estimand & Separate 72 & 504-record panel \\
\midrule
$\beta_O$ & $0.214\ [0.174,0.258]$ & $0.261\ [-0.129,0.785]$ \\
$\beta_R$ & $0.214\ [0.175,0.255]$ & $0.218\ [0.023,0.510]$ \\
$A$ & $0.0003\ [-0.0157,0.0166]$ & $0.043\ [-0.136,0.254]$ \\
$U_V$ & $0.040\ [0.024,0.056]$ & $0.027\ [0.016,0.038]$ \\
$R_O$ & $0.014\ [-0.003,0.032]$ & $0.010\ [0.002,0.017]$ \\
$R_R$ & $0.011\ [-0.007,0.030]$ & $0.012\ [0.004,0.019]$ \\
\bottomrule
\end{tabular}
\caption{Estimates with 95\% CIs from two non-overlapping panels. The separate panel applies one corrected protocol to all seven targets in 72 records. The 504-record panel uses all records for two unaffected targets and corrected forecasts from 72 within-panel records for five affected targets, with equal target weight.}
\label{tab:core-results}
\end{table}

\begin{table*}[!t]
\centering
\small
\setlength{\tabcolsep}{5.0pt}
\begin{tabular}{@{}lccc@{}}
\toprule
\multicolumn{4}{@{}l}{\textit{Panel A. Country-directed movement and disclosure attenuation}} \\
API model & $\beta_O$ [95\% CI] & $\beta_R$ [95\% CI] & $A$ [95\% CI] \\
\midrule
DeepSeek-V4-Pro & $0.094\ [0.019,0.183]$ & $0.093\ [0.027,0.174]$ & $0.001\ [-0.041,0.039]$ \\
Gemini-3.6-Flash & $0.397\ [0.330,0.461]$ & $0.363\ [0.287,0.433]$ & $0.035\ [-0.004,0.080]$ \\
Mistral-Medium-3.5 & $0.122\ [0.071,0.180]$ & $0.142\ [0.087,0.195]$ & $-0.020\ [-0.048,0.015]$ \\
GPT-5.4 & $0.152\ [0.099,0.199]$ & $0.147\ [0.100,0.194]$ & $0.005\ [-0.029,0.033]$ \\
Qwen3.7-Max-2026-06-08 & $0.305\ [0.226,0.396]$ & $0.323\ [0.242,0.412]$ & $-0.019\ [-0.058,0.028]$ \\
\midrule
\multicolumn{4}{@{}l}{\textit{Panel B. Held-out predictive consequences}} \\
API model & $R_O$ [95\% CI] & $R_R$ [95\% CI] & $U_V$ [95\% CI] \\
\midrule
DeepSeek-V4-Pro & $-0.014\ [-0.037,0.008]$ & $-0.013\ [-0.037,0.012]$ & $0.042\ [0.017,0.067]$ \\
Gemini-3.6-Flash & $0.016\ [-0.015,0.054]$ & $0.006\ [-0.026,0.043]$ & $0.061\ [0.038,0.085]$ \\
Mistral-Medium-3.5 & $0.007\ [-0.013,0.026]$ & $0.011\ [-0.010,0.034]$ & $0.033\ [0.017,0.052]$ \\
GPT-5.4 & $0.009\ [-0.012,0.027]$ & $0.004\ [-0.020,0.030]$ & $0.034\ [0.015,0.055]$ \\
Qwen3.7-Max-2026-06-08 & $0.052\ [0.024,0.082]$ & $0.047\ [0.016,0.078]$ & $0.027\ [0.002,0.055]$ \\
\bottomrule
\end{tabular}

\begin{tabular}{@{}lccccccc@{}}
\multicolumn{8}{@{}l}{\textit{Panel C. Target-level direction and consequence (descriptive)}} \\
Metric & A124\_09 & E025 & A124\_08 & A124\_03 & A030 & E027 & A029 \\
\midrule
$\beta_R$ & $0.283$ & $0.309$ & $0.201$ & $0.238$ & $0.096$ & $0.305$ & $0.063$ \\
$A$ & $0.018$ & $-0.015$ & $0.025$ & $0.016$ & $0.016$ & $-0.017$ & $-0.040$ \\
$R_R$ & $0.032$ & $0.055$ & $-0.012$ & $0.012$ & $-0.029$ & $0.014$ & $0.004$ \\
$U_V$ & $0.144$ & $0.012$ & $0.052$ & $0.028$ & $0.025$ & $0.022$ & $-0.006$ \\
\bottomrule
\end{tabular}
\caption{Model- and target-level results in the separate 72-record panel. All model-level intervals use 2,000 draws under the corresponding clustered direction or paired-Brier scheme and are descriptive and unadjusted. Panel C reports every target's post-review exploratory point estimate without item-wise inference.}
\label{tab:model-results}
\end{table*}

\subsection{Statistical Inference}

Separate-panel direction intervals use 10,000 draws that resample evaluation records within survey country and independently resample reference records to rebuild the anchors. Brier intervals use 10,000 paired within-country draws. Country, item, and model cells receive equal macro weight. RQ1 requires both $\beta_O>0$ and $A>0$. RQ2 tests $U_V>0$, with Holm adjustment across the two questions. A 100,000-draw balanced-label permutation tests the stronger Fisher sharp null for direction. Reference magnitudes are 0.05 attenuation and 0.02 verified utility. The 504-record panel uses 10,000 nested record and reference bootstrap draws. Its item-specific coverage precludes a second test under the separate-panel specification.

Descriptive model-level, target-level, and leave-one-target-out intervals use 2,000 draws with the corresponding direction or paired-Brier resampling scheme. They are not multiplicity-adjusted, and the leave-one-target-out analysis assesses concentration rather than generalisation to new targets.

\section{Results}

\paragraph{Random-label uptake and provenance disclosure.}
Both random-label conditions produced positive country-directed movement (Table~\ref{tab:core-results}). Their paired difference was near zero, and the RQ1 intersection-union test gave $p = 0.498$. The balanced-label Fisher sharp-null sensitivity gave the same decision ($p = 0.482$). The interval for $A$ excludes attenuation as large as the 0.05 reference magnitude, but it includes zero and smaller effects of either sign.

\paragraph{Verified-country utility.}
The verified-country policy lowered held-out Brier loss on the seven selected targets. The Holm-adjusted one-sided RQ2 test remained below 0.001. The lower-bound test against the 0.02 utility reference magnitude gave $p = 0.008$.

\paragraph{Mixed-coverage consistency estimates.}
The 504-record panel is a secondary consistency analysis, not a complete seven-target replication. It excluded every affected old-wording forecast. Disclosed-random direction was $0.218$ $[0.023,0.510]$, attenuation was $0.043$ $[-0.136,0.254]$, and verified utility was $0.027$ $[0.016,0.038]$ (Table~\ref{tab:core-results}). Movement after disclosure and verified utility were positive in both panels, while attenuation remained uncertain. Both mixed-coverage-panel regret intervals excluded zero, whereas both separate-panel intervals crossed zero.

\paragraph{Model-level estimates.}
Panels A and B of Table~\ref{tab:model-results} show positive opaque-label movement for all five models, while every model-specific attenuation interval includes zero. Verified-country utility intervals were positive for all five models; disclosed-random regret intervals excluded zero only for Qwen3.7-Max-2026-06-08. These intervals are descriptive and not adjusted for model-level multiplicity. The supplementary material reports the post-hoc model-by-country analysis and complete matrices.

\paragraph{Target-level concentration.}
Panel C reports estimates for all seven targets. Disclosed-random direction point estimates ranged from $0.063$ to $0.309$; attenuation ranged from $-0.040$ to $0.025$, with all seven pointwise intervals including zero. Disclosed-random regret point estimates ranged from $-0.029$ to $0.055$, and verified-country utility from $-0.006$ to $0.144$. The smallest descriptive leave-one-target-out utility estimate was $0.022$ (pointwise 95\% CI $[0.005,0.040]$) when A124\_09 was omitted.

\section{Discussion and Limitations}

The joint result is a behavioural asymmetry under the tested protocol. Survey country helped when copied from the source record on targets selected for incremental country value. Explicitly stating that the same displayed label was random and independent did not reliably reduce its country-directed effect. This result is narrower than a claim that the models cannot understand provenance. It evaluates one disclosure, fixed English wording, five-question batches, and probability forecasts.

Direction and consequence also diverged. A forecast can move towards a displayed country's population contrast without incurring detectable average Brier harm. Because direction and utility use different scales, their coexistence does not establish miscalibration or identify a mechanism. Training-data associations, instruction following, salience, and probability elicitation remain observationally entangled.

At target level, disclosed-random direction point estimates were positive throughout, whereas regret point estimates spanned both signs and verified utility was uneven. This is a distributional diagnosis rather than evidence that every target exhibits the same effect.

Several boundaries constrain generalisation. The 504-record panel is a mixed-coverage consistency analysis because only two targets use all 504 records; five affected targets use corrected forecasts from 72 records within that panel. The separate 72-record panel covers all seven targets but was not prospectively powered. Verified-country utility therefore concerns development-selected high-country-value targets rather than survey questions in general. In separate report-only diagnostic panels, mean forecast TV was $0.105$ under single-item presentation, $0.120$ after option reversal, and $0.117$ when companion questions changed. An identical-prompt repeat panel yielded $0.041$. These auxiliary panels used different record and target subsets and did not re-estimate the current corrected-panel estimands. They limit claims to the frozen elicitation contract and do not establish interface-invariant effects. Survey country is an administrative collection field, not nationality or cultural essence. Harmonised English prompts differ from the local-language survey contexts, and cross-national measurement equivalence is not established. The design elicits output distributions rather than internal beliefs or calibrated confidence and cannot isolate truth status within $M_V$ because the complete verified-metadata policy changes. It also audits a randomised cue effect rather than a causal fairness criterion. Future work should prospectively randomise disclosure wording, batch composition, and option order and re-estimate the primary distributional estimands across survey waves and fixed model releases.

\section{Conclusion}

Across five fixed API models, verified survey-country metadata lowered held-out loss on the selected targets, while disclosing a random label's record-independent origin did not reliably attenuate its country-directed uptake. The primary uniform panel and secondary mixed-coverage analysis retained verified utility and disclosed-random movement, but did not support a stable conclusion about average random-label harm. PROV-FORECAST provides paired distributions for studying metadata provenance, probabilistic cue uptake, and cross-model heterogeneity within the released protocol. Future work should test varied provenance disclosures and unbatched forecasts to determine whether smaller attenuation effects generalise beyond the present models and survey setting.

\section*{Ethical Statement}

The study reuses existing pseudonymous survey records and recruits no participants. Prompts omit source identifiers, survey weights, reference anchors, and held-out answers. True survey country appears only in $M_V$. Respondent-level source data and record-derived prompts are not redistributed. The reproducibility artifact releases sanitized item-level model forecasts under release-specific paired-unit identifiers, but excludes observed and held-out survey answers, raw response text, provider request identifiers, and timing or token metadata. It also contains code, item mappings, prompt templates, frozen manifests, hashes, and aggregate outputs, and requires users to obtain the Joint EVS/WVS data under its access conditions to reconstruct respondent-level scoring. CC BY 4.0 covers the released forecast table and data card; no broader source-data rights are granted. Public non-commercial research access terms do not expressly address third-party hosted inference. No endpoint-specific authorisation from EVS/WVS or GESIS for such processing is claimed, which limits the governance claim to the implemented pseudonymisation, minimisation, and non-redistribution controls.

\bibliography{references}

\end{document}